# Enhancing Medical Specialty Assignment to Patients using NLP Techniques

CHRIS SOLOMOU

University of York - Department of Computer Science

Deramore Lane Heslington York, York UK

The introduction of Large Language Models (LLMs), and the vast volume of publicly available medical data, amplified the application of NLP to the medical domain. However, LLMs are pretrained on data that are not explicitly relevant to the domain that are applied to and are often biased towards the original data they were pretrained upon. Even when pretrained on domain-specific data, these models typically require time-consuming fine-tuning to achieve good performance for a specific task. To address these limitations, we propose an alternative approach that achieves superior performance while being computationally efficient. Specifically, we utilize keywords to train a deep learning architecture that outperforms a language model pretrained on a large corpus of text. Our proposal does not require pretraining nor fine-tuning and can be applied directly to a specific setting for performing multi-label classification. Our objective is to automatically assign a new patient to the specialty of the medical professional they require, using a dataset that contains medical transcriptions and relevant keywords. To this end, we fine-tune the PubMedBERT model on this dataset, which serves as the baseline for our experiments. We then twice train/fine-tune a DNN and the RoBERTa language model, using both the keywords and the full transcriptions as input. We compare the performance of these approaches using relevant metrics. Our results demonstrate that utilizing keywords for text classification significantly improves classification performance, for both a basic DL architecture and a large language model. Our approach represents a promising and efficient alternative to traditional methods for fine-tuning language models on domain-specific data and has potential applications in various medical domains.

## 1 INTRODUCTION

A Medical Specialty refers to a field of medicine that defines a group of diseases, patients, and the professionals that treat them. Each specialty requires its own set of skills, treatments, and medicine for dealing with specific health conditions. Relevant to the work of this paper, each medical specialty has specific keywords that distinguish it from the rest. This, in turn, allows for the applications of modern NLP algorithms to be applied for predicting the specialty of the medical professional that a newly admitted patient needs.

Communication between medical professionals and patients is an important process in medical care. In reviewing these communications, topics such as the purpose of the medical communication and the preliminary health condition of the patient can be identified [1]. These data, if recorded, can offer a lot of value if used to train ML models which can be used for automatically predicting the condition of a patient, and the medical care that he/she needs.

Generally, applications of ML algorithms in the medical domain offer an immediate and unbiased diagnosis since they can be trained on past data of patients with similar symptoms. By being able to predict the medical specialty that a patient needs, a medical practitioner can make a more informed diagnosis regarding the condition of a patient, and how he/she should be treated [3]. Moreover, by providing an immediate diagnosis, the patient can avoid visiting a medical practitioner who is irrelevant to the underlying condition, thus saving time that could prove detrimental to the patient's recovery. The above statement motivates a lot of applications of Computer Vision in the healthcare domain, with applications in medical imaging including using x-rays to predict the age, gender [3], and health condition of a patient [12,13,14]. Specifically, when trained on large volumes of data these models can produce similar results to domain experts for the diagnosis and treatment of diseases [2].

Similarly, the introduction of large language models and the abundance of publicly available medical data from sources such as PubMed, have boosted the application of NLP in healthcare. However, they suffer from the fact that are pretrained on data that might not be explicitly relevant to the area of interest, (i.e., Wikipedia articles). Even when these models are pretrained specifically on data from the domain in question, they still have to be fine-tuned to perform a specific application. These operations are often time-consuming, computationally expensive, and do not guarantee good results.

In this work, we overcome these issues with the framework we propose, applied to multi-label classification tasks (i.e., predicting the specialty of the medical professional depending on patients' health condition). It is usually the case that when a patient visits a hospital a medical professional transcribes in a keyword-style the symptoms of the patient. Our proposal illustrates that utilizing these keywords as input to a DL model constitutes the primary mechanism for attaining state-of-the-art performance while keeping computational costs low, with the model selection assuming a secondary role.

In subsequent parts of this paper, we review past work of ML models applied to healthcare and give the background of how our proposal compares with existing solutions. Subsequently, we compare our proposal with a state-of-the-art language model pretrained on the abstracts from PubMed, and another language model pretrained on a general corpus of text. Our aim is to perform multi-label classification in the context of assigning the specialty of a medical professional to a new patient based on the patient's symptoms. We present the results of our efforts and discuss the best practices and limitations of our approach.

## 2 RELATED WORK

There is a rich history of the application of ML models to textual medical data. Wallace et al. [4] proposed a Conditional Random Field (CRF) to model the topic probabilities of patient-provider communication, achieving above-random results in predicting the topic of each sentence. Their work emphasized the importance of automated topic annotation to help medical providers focus on relevant topics.

Zhong et al. [8] created patients' clinical profiles from unstructured clinical records in Chinese and utilized these profiles to predict each patient's disease category based on the ICD-10 coding system. Through experimentation with different Deep Learning and traditional ML algorithms, they discovered that convolutional neural networks (CNNs) outperformed other approaches in terms of F1-score.

Similarly, Seva et al. [10] used a language-independent neural architecture to extract death causes from death certificates according to the ICD-10 codes. Their approach consisted of two recurrent neural networks, one for extracting the death-cause text and the other for assigning the respective code of death. Their work yielded promising results with their methodology being applicable to other languages as well.

On the other hand, the revolution of transformer models in NLP such as BERT [5], resulted in excellent performance for extracting meaning from unstructured data. Inspired by the success of transformers, numerous attempts have been made to apply these models to other, more specific domains. However, these models are already pretrained in large corpuses of unspecific text, that do not yield the same results when applied in the healthcare domain.

Gu et.al. [15] proposed that general-domain language models perform much worse than similar models pretrained on a specific domain. The authors pretrained from scratch a BERT architecture using the abstracts from PubMed articles and showed that such a model can achieve state-of-the-art performance in a variety of tasks such as named entity recognition, relation extraction, document classification, and question answering.
Other adoptions of domain-specific transformer models include BioBert [6] and DiLBert [7]. Specifically, BioBERT demonstrates superior performance over other state-of-the-art models in biomedical named entity recognition, relation extraction, and question answering. Furthermore, by pretraining BERT on a biomedical corpus enhances its ability to comprehend complex biomedical texts. Delvin et.al [5], trained a BERT architecture using the original text from English Wikipedia, Books Corpus, and PubMed, demonstrating that having domain-specific knowledge can significantly outperform similar state-of-the-art models that were trained in a general domain.

Although directly training on a specific domain can improve model performance, there is still further room for specialization. For instance, accurately interpreting expressions related to diseases and diagnoses requires specialized models to achieve optimal results. To this end, Roitero et al. [7], who developed a more specific model by constructing a database of ICD-11 entities and integrating relevant data from Wikipedia and PubMed. This approach led to the development of DiLBert, a language model that achieves state-of-the-art performance using a much smaller dataset.

There are also similar approaches where the authors utilized large language models, pretrained solely on electronic health records. Rasmy et.al. [16] using the BERT framework, proposed Med-BERT, a language model pretrained on structured electronic health records from Cerner Health Facts. The model managed to outperform other popular models in the task of predicting heart failure in patients with diabetes.
Despite the successes of NLP models, there is still room for specialization in interpreting expressions mentioning diseases and diagnoses, where specialized models can accomplish specific tasks even better.

However, the current research is only concerned on going deeper (more relevant data) and bigger (increasing the size of pretraining data) rather than experimenting with different techniques (i.e., preprocessing) for improving performance.

## 3 BACKGROUND

Our work is based on the proposal by [15], namely (pre)training a model on data from a specific domain improves the performance of the task at hand. Taking this idea further, we claim that training a model directly on specific keywords yields even superior performance compared to using an in-domain text corpus. Therefore, we demonstrate that the characteristics of the model become less important, with the quality and nature of the data being the primary factors influencing model performance.

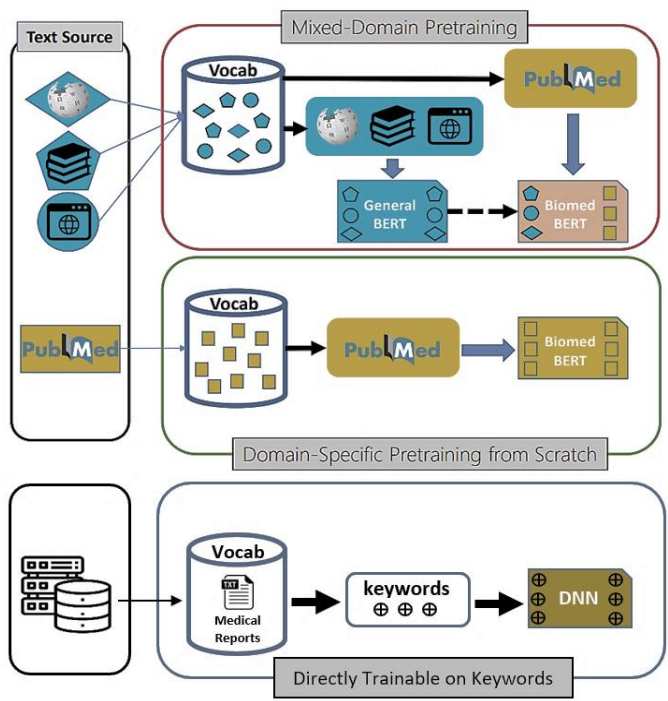

**Figure 1. Three paradigms of NLP models applied to the medical domain. The first two paradigms as presented by [15] and our proposed approach on the bottom. The first paradigm assumes that including general text data alongside medical data enhances the overall performance of the model, while the second paradigm proposes pretraining the model exclusively on medical data. Our proposal leverages medical data containing keywords for direct model training, eliminating the need for a separate pretraining step (and subsequent fine-tuning), thus avoiding unnecessary computational costs altogether.**

For this purpose, a dataset that includes medical transcriptions and their respective keywords for describing the health condition of a patient is utilized. These data are used as input for training NLP models that can be used to make an informed diagnosis about the medical specialty that a newly admitted patient requires. This process can eliminate the need for computationally expensive pretraining and fine-tuning of language models, leading to faster and more accurate diagnoses.

To test the effectiveness of our approach, we compare the performance of PubMedBERT [15], a language model pretrained on medical text, RoBERTa [9], a general pretrained language model, and our proposed model using both full transcriptions and keywords. We evaluate the models using the metrics described in the next section.

## 4 MATERIALS AND METHODOLOGY

In this section, we provide an overview of the methodology employed in our work. We discuss the model architecture, dataset, preprocessing techniques, and evaluation metrics used to assess the performance of each approach.

**4.1 Dataset:** Our training dataset is composed of 4999 medical transcriptions scraped from mtsamples.com [11], which include six features. We use the medical specialty feature as the target class for our models. The transcription and keywords features are used to train each model. To ensure reliable performance estimates, we employ a stratified 5-fold cross-validator that preserves the percentage of samples for each class. Oversampling or undersampling techniques like SMOTE are not effective for heavily imbalanced datasets, and can introduce biases in the model's training process, leading to poor performance on underrepresented classes. In contrast, k-fold cross-validation provides a more robust assessment of model performance by ensuring balanced representation of each class across different folds, reducing bias, and enabling better evaluation on imbalanced data.

| Medical Specialty | Occurrences | Medical Specialty | Occurrences |
|---|---|---|---|
| Surgery | 1103 | Psychiatry / Psychology | 53 |
| Consult - History and Physio | 516 | Office Notes | 51 |
| Cardiovascular / Pulmonary | 372 | Podiatry | 47 |
| Orthopedic | 355 | Dermatology | 29 |
| Radiology | 273 | Dentistry | 27 |
| General Medicine | 259 | Cosmetic / Plastic Surgery | 27 |
| Gastroenterology | 230 | Letters | 23 |
| Neurology | 223 | Physical Medicine - Rehab | 21 |
| SOAP / Chart / Progress Notes | 166 | Sleep Medicine | 20 |
| Obstetrics / Gynecology | 160 | Endocrinology | 19 |
| Urology | 158 | Bariatrics | 18 |
| Discharge Summary | 108 | IME-QME-Work Comp etc. | 16 |
| ENT - Otolaryngology | 98 | Chiropractic | 14 |
| Neurosurgery | 94 | Diets and Nutrition | 10 |
| Hematology - Oncology | 90 | Rheumatology | 10 |
| Ophthalmology | 83 | Speech - Language | 9 |
| Nephrology | 81 | Autopsy | 8 |
| Emergency Room Reports | 75 | Lab Medicine - Pathology | 8 |
| Pediatrics - Neonatal | 70 | Allergy / Immunology | 7 |
| Pain Management | 62 | Hospice - Palliative Care | 6 |

**Table. 1: Distribution of Class Labels**

Additionally, by evaluating the models on this dataset we aim to test their ability to handle challenging real-world situations where certain diseases are more prevalent than others. This approach presents a greater challenge to the models and allows us to evaluate their performance using relevant metrics that specifically account for class-imbalance.

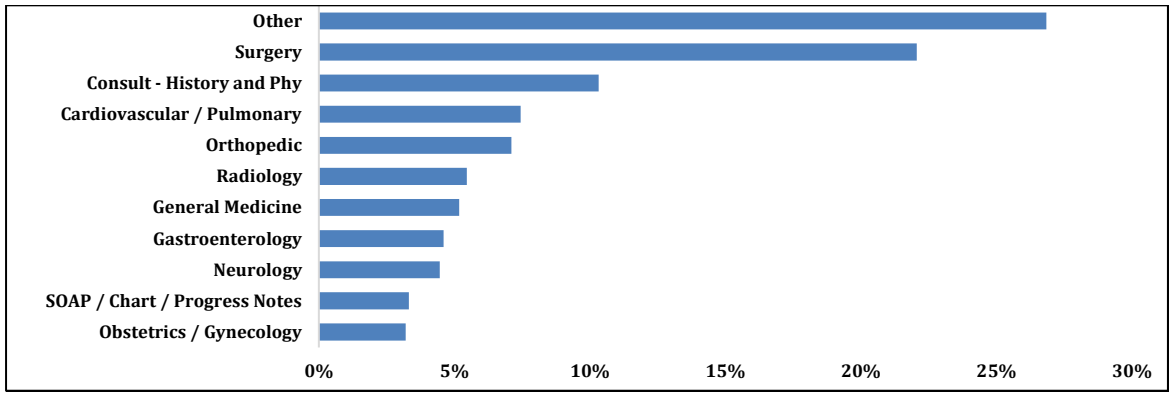

**Figure 2:** The proportion of the top 10 class-labels compared with the rest (Other). This presents a difficult classification task due to the presence of numerous classes, many containing a limited number of instances.

### 4.2 Medical specialty classification with a DNN trained directly on medical keywords:

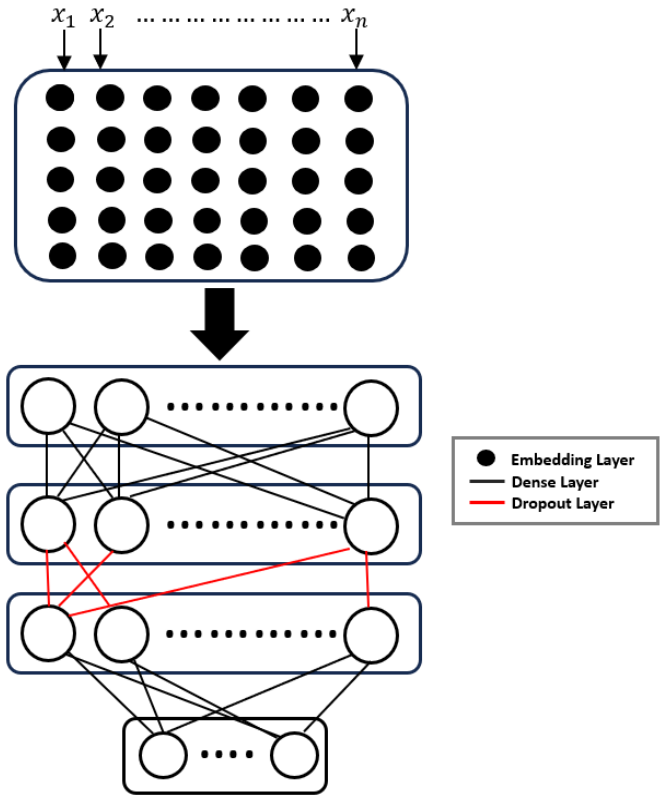

This basic architecture comprises a DNN that sits on top of an embedding layer. Thereafter, a fully connected layer is added, with Batch Normalization applied to its output values. Subsequently, another two fully connected layers are added, consisting of 100 neurons each. Relu activation functions are applied to the fully connected layers, except the output layer which utilizes a softmax activation function for multi-class output.

The model presented in this section will be trained twice. Once for predicting the class of the medical specialty using keywords, and once using the medical transcriptions as input data. The preprocessing steps for both approaches are to remove missing records, convert the text to lowercase and remove punctuations. Additionally, we omit stopwords (i.e., me, the, he, etc.) that offer little value for predicting a specific class. For performing classification with the medical transcriptions, the records were truncated at 120 words, and 15 words when using keywords, with padding also applied. The data were split into 5 folds with the same random seed, so all models are trained/fine-tuned and evaluated using the same data set. In both cases, the model was trained for 200 epochs, or with early stopping if no improvement occurred after 10 epochs.

**4.3 Medical specialty classification with fine-tuning pretrained large language models (LLMs):**
As a comparison to our proposal, two large language models will be employed, RoBERTa [9] which was pretrained on a text corpus that is not relevant to the medical domain, and PubMedBERT [15] which was pretrained on the abstracts from PubMedBERT. When fine-tuning the models, the data were preprocessed using the respective tokenizer of each model. An example of preprocessing and fine-tuning is illustrated in Fig 4, where the tokenizer generates the input ids and attention masks, based on the model's vocabulary.

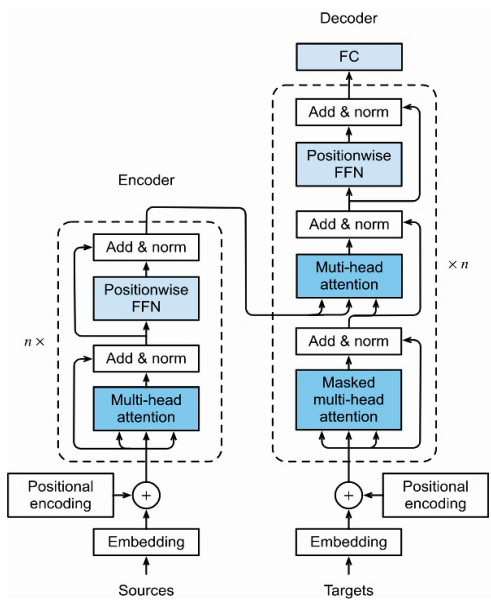
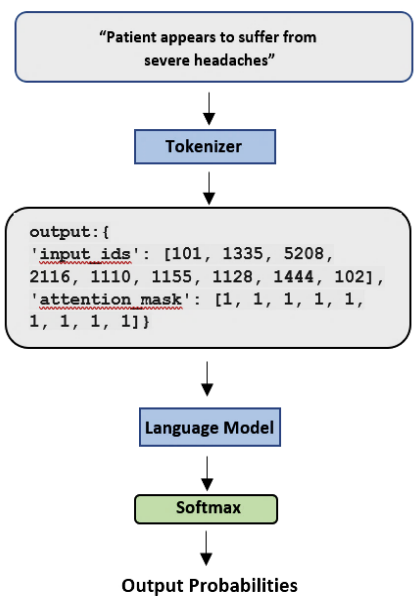

**Figure 4: Architecture of BERT language model**    **Figure 5: Fine-tuning on new data**

For performing classification with the medical transcriptions, the records were truncated at 120 words with padding also applied. Both models were fine-tuned for 20 epochs. For performing classification using keywords, the records were truncated at 15 words as 4.2 and fine-tuned for 20 epochs, with early stopping if the model did not improve after 3 epochs. The dataset was split the same way as described in 4.2 with the same random seed, so all models face the same data. As before, missing records were removed. To facilitate a fair comparison between the two language models, we apply the same fine-tuning procedure. In both models the Adam optimizer was used with the same learning rate as the original papers, decaying at a constant rate.

**5 EXPERIMENTS AND RESULTS**
In this section we present the results of training/fine-tuning on keywords and the full medical transcriptions. The evaluation metrics employed to assess the success of each model and training approach include Precision, Recall and F1 score. The F1 score provides a comprehensive assessment of model performance by combining precision and recall into a single metric. The results are averaged for each performance metric across the 5 folds that were used for training and evaluating the models.

| PubMedBERT | Keywords | | | Transcriptions | | |
|---|---|---|---|---|---|---|
| Metric | Precision | Recall | F1-score | Precision | Recall | F1-score |
| Accuracy (micro avg) | | | 0.54 ± 0.31 | | | 0.30 ± 0.07 |
| Macro avg | 0.38 ± 0.45 | 0.40 ± 0.45 | 0.39 ± 0.45 | 0.02 ± 0.02 | 0.05 ± 0.02 | 0.03 ± 0.02 |
| Weighted avg | 0.45 ± 0.39 | 0.54 ± 0.31 | 0.46 ± 0.37 | 0.11 ± 0.06 | 0.30 ± 0.07 | 0.15 ± 0.06 |

Table 2: Results of PubMedBERT fine-tuned on keywords and transcriptions.

Applying the language models on a small and imbalanced dataset uncovers their limitations. Both models appear to suffer from some inherited bias deriving from the datasets that were originally pretrained upon. Hence, when fine-tuned using the full transcriptions, both models perform poorly. PubMedBERT appears to perform slightly better than RoBERTa having the advantage of being pretrained on data related to healthcare. However, when fine-tuned on keywords, both models perform much better, achieving improved performance considering the challenging nature of the dataset. It is noteworthy that although the predictive capabilities of the models have significantly improved, the error of the predictions became also larger.

| RoBERTa | Keywords | | | Transcriptions | | |
|---|---|---|---|---|---|---|
| Metric | Precision | Recall | F1-score | Precision | Recall | F1-score |
| Accuracy (micro avg) | | | 0.56 ± 0.30 | | | 0.25 ± 0.06 |
| Macro avg | 0.40 ± 0.46 | 0.41 ± 0.44 | 0.40 ± 0.46 | 0.01 ± 0.02 | 0.04 ± 0.02 | 0.02 ± 0.02 |
| Weighted avg | 0.47 ± 0.34 | 0.56 ± 0.30 | 0.49 ± 0.34 | 0.08 ± 0.06 | 0.25 ± 0.05 | 0.11 ± 0.06 |

Table 3: Results of RoBERTa fine-tuned on keywords and transcriptions.

The DNN on the other hand performs worst compared to the language models when trained on the medical transcriptions. Even though we removed the stopwords that do not add value to the model, it was not enough to make the relevant words stand out. Nevertheless, when we use keywords to train our model, the opposite occurs, with a vast increase in performance. The predictive capability of the DNN far exceeds that of the large language models, with the error of the predictions significantly reduced.

| DNN | Keywords | | | Transcriptions | | |
|---|---|---|---|---|---|---|
| Metric | Precision | Recall | F1-score | Precision | Recall | F1-score |
| Accuracy (micro avg) | | | 0.81 ± 0.01 | | | 0.18 ± 0.01 |
| Macro avg | 0.84 ± 0.02 | 0.66 ± 0.02 | 0.72 ± 0.01 | 0.08 ± 0.01 | 0.07 ± 0.00 | 0.07 ± 0.00 |
| Weighted avg | 0.90 ± 0.00 | 0.81 ± 0.01 | 0.83 ± 0.01 | 0.16 ± 0.06 | 0.18 ± 0.01 | 0.17 ± 0.01 |

Table 4: Results of DNN fine-tuned on keywords and transcriptions.

Generally, all three models do not perform well for predicting the medical specialty using the full transcriptions as input. We believe this primarily happens because of the nature of the dataset, i.e., being small and imbalanced. Specifically for the language models, being pretrained on very large volumes of text, prevents them from being specific to the current application. Especially in the case of RoBERTa, since PubMedBERT performed marginally better having the benefit of being pretrained on medical data. Conversely, when the language models are fine-tuned on keywords, they perform much better, since they prevent the models from generalizing to other non-relevant words. This results to better performance, although the results between the two models are quite similar.The DNN performs better when trained on the present data, avoiding bias towards an original dataset. However, training on the full medical transcriptions results in poor performance due to relevant words being masked by the rest. On the contrary, training on keywords allows the model to focus on applicable words and capture relationships more effectively. The DNN is also computationally efficient and can be trained for longer periods than the large language models if necessary.

In summary, we empirically show that using keywords as input to a basic DNN produces better results than large pretrained language models with much less computational cost. We also show that in cases where the language models do not perform well, their performance can significantly be improved by using keywords for fine-tuning.

**6 DISCUSSION AND FUTURE WORK**

In this work, we propose a new approach for automatically predicting the medical specialty that a patient necessitates. Our proposal of using keywords to assign a new patient to a medical specialty, offers superior accuracy, with much less computational cost. Unlike other large language models that must be pretrained on a specific domain, our proposal learns directly from the data of the current application. This makes for a practical alternative since training directly on the data, makes obsolete the need to pretrain, and later fine-tune when the domain of interest changes. Nonetheless, language models with no specific domain knowledge, can also benefit from this approach, resulting in improved performance.

While many text documents typically lack the necessary keywords, this specific scenario fulfills the keyword requirement through the symptoms communicated by a patient during their initial visit to a medical facility. Alternatively, these keywords could be engineered to suit our proposal, though we leave this aspect for future exploration. Consequently, the symptom-keywords, which describe the patient's condition, can be effectively utilized to train a range of deep learning architectures that achieve superior performance compared to training on complete medical reports, where such models might falter.

Our findings are highly promising, particularly given the challenging nature of the dataset. As larger datasets are incorporated to train these models, we anticipate achieving near-perfect accuracy. This, in turn, will yield numerous benefits, including immediate care for patients, validation of diagnoses and shorter recovery times.

The work performed in this paper sets the foundation for future work that we aim to carry out. Specifically, most medical reports contain valuable demographic characteristics of the patients, such as age, gender, and medical history. Adjusting the same methodology, we can recommend appropriate medical specialties by taking these factors into account. Moreover, since patients' symptoms may fall under multiple medical specialties, we need to consider an architecture capable of predicting multiple labels and classes. Furthermore, we aim to enhance the proposed method's robustness by leveraging existing medical reports to extract new keywords. Utilizing this vast repository of past medical reports, while ensuring anonymity to adhere to privacy regulations, will bolster the effectiveness of our approach.